\documentclass[conference]{IEEEtran}
\usepackage{cite}
\usepackage{hyperref}

\usepackage[utf8]{inputenc}
\usepackage{listings}
\usepackage{rotating}
\usepackage{siunitx}
\usepackage{amsmath,amssymb,amsfonts}
\usepackage{algorithmic}
\usepackage{graphicx}
\usepackage{textcomp}
\usepackage{xcolor}
\usepackage{booktabs}

\usepackage{amsthm}

\newtheorem*{remark}{Remark}
\newtheorem{condition}{Condition}

\graphicspath{ {./figures/} }

\def\BibTeX{{\rm B\kern-.05em{\sc i\kern-.025em b}\kern-.08em
    T\kern-.1667em\lower.7ex\hbox{E}\kern-.125emX}}

\IEEEoverridecommandlockouts
\begin{document}

\title{Statistical Modelling of Level Difficulty in Puzzle Games}

\author{\IEEEauthorblockN{Jeppe Theiss Kristensen}
\IEEEauthorblockA{ 
\textit{IT University of Copenhagen/Tactile Games}\\
Copenhagen, Denmark \\
jetk@itu.dk}
\and
\IEEEauthorblockN{Arturo Valdivia*}\thanks{* Corresponding author.}
\IEEEauthorblockA{ 
\textit{Tactile Games}\\
Copenhagen, Denmark \\
arturo@valdivia.xyz}
\and
\IEEEauthorblockN{Paolo Burelli}
\IEEEauthorblockA{
\textit{IT University of Copenhagen/Tactile Games}\\
Copenhagen, Denmark \\
pabu@itu.dk}
}


\maketitle

\begin{abstract}

Successful and accurate modelling of level difficulty is a fundamental component of the operationalisation of player experience as difficulty is one of the most important and commonly used signals for content design and adaptation.
In games that feature intermediate milestones, such as completable areas or levels, difficulty is often defined by the probability of completion or completion rate; however, this operationalisation is limited in that it does not describe the behaviour of the player within the area.

In this research work, we formalise a model of level difficulty for puzzle games that goes beyond the classical probability of success.
We accomplish this by describing the distribution of actions performed within a game level using a parametric statistical model thus creating a richer descriptor of difficulty.
The model is fitted and evaluated on a dataset collected from the game Lily's Garden by Tactile Games, and the results of the evaluation show that the it is able to describe and explain difficulty in a vast majority of the levels.

\end{abstract}

\begin{IEEEkeywords}
player modelling, difficulty modelling, game design, dda, survival analysis
\end{IEEEkeywords}

\section{Introduction}
A central aspect of game design is difficulty and its effect on player experience -- too easy and players are not sufficiently engaged; too hard and players become frustrated, causing them to quit the game.
In games consisting of discrete tasks or levels, a common way to manage the difficulty is by controlling the resources available to the player to complete such task or level -- \emph{e.g.}, number of actions or time available to solve a puzzle.
Balancing the correct number of resources available in the level to obtain a desired difficulty is a complex task that often relies on the ability of the designer to relate an abstract descriptor of difficulty to the behaviour of the players and the controllable components in the level.

For example, in the case of puzzle games that provide players with limited actions, or \textit{moves}, to complete each level, such as match-3 or bubble shooter style games, a direct way to describe the difficulty is by measuring how many attempts it takes players on average to complete a level.
This quantity is commonly referred to as \textit{attempts-to-complete}, and its multiplicative inverse is what we call \textit{completion rate}.
This definition is useful for identifying levels in which players may feel stuck and thus stop playing, controlling the consumption rate of game content, or even enabling different monetisation strategies.
However, such descriptor only considers the data in an aggregated way and thus lacks the granularity that may, for example, tell about the effect of changing the action limit or how close to finish a player was.
This makes it relatively limited in it expressiveness, giving a designer little information on how to adjust the difficulty and thus turning the task of level adjustment into a trial-and-error procedure.

In the vast majority of currently published puzzle games, success or failure are not the only data available about the player behaviour in a game; often a summary of the actions performed and the resources used are tracked.
If properly modelled, this information has the potential to be the basis of a much richer descriptor of level difficulty. 
In particular, the number of actions used by players in their attempts has both the benefit of describing their progress within a level and being directly related to an important level design aspect, move limit.

\begin{figure}[t]
    \centering
    \includegraphics[trim={0 2 0 2},clip,width=0.82\columnwidth]{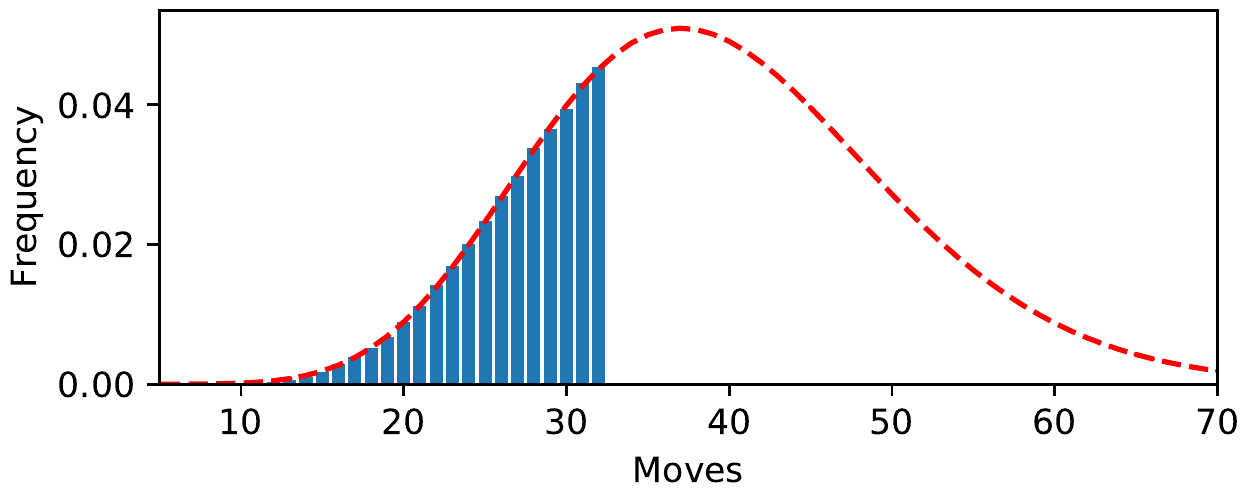}
    \caption{Histogram over the number of actions spent by players to complete one of the levels in the data. The effect of the action limit near $M=32$ can clearly be seen as a sharp cut-off in the distribution. If we are able to accurately estimate the full distribution (represented by the red curve), the completion rate using different action limits can be calculated.}
    \label{fig:example move distribution}
\end{figure}

The number of actions used to complete a level depend on a number of factors, such as player skill, level setup and luck.
This leads to a certain distribution of actions spent by players on each level (see Fig. \ref{fig:example move distribution}).
The central idea of this article is that, by modelling and understanding the nature of this action distribution, we may be able to not only evaluate the completion rate but also estimate the effect of design actions, such as changing the move limit, and gain a deeper understanding of the player challenges. 

To achieve this, the model of the player behaviour needs to be both accurate and explainable. 
For this reason, in this research work, we have investigated the application of a parametric statistical model to represent the underlying action distribution.
We discuss how this behaviour can be modelled using a negative binomial distribution and conduct an empirical study of the application of this modelling approach to a dataset from a popular mobile puzzle game -- Lily's Garden by Tactile Games -- and present and discuss the results of the study.



\section{Related work}

\textit{Flow} \cite{csikszentmihalyi1990flow} describes the psychological state where the difficulty of a task and user skill match which leads to an engaging gameplay experience.
While difficulty can be broken down into multiple sub-components (e.g. cognitive, emotional, etc. \cite{DENISOVA2020102383}), in scenarios where it is necessary to operationalise difficulty, such as for dynamic difficulty adjustment or automated playtesting, it is common to use the probability of task success as an objective measure of difficulty \cite{Lomas2017IsDifficultyOverrated, xue2017dynamic,Duque2020FindingTrial-and-Error,Demediuk2017MonteAdjustment, kristensen2020estimating, gudmundsson2018human}.
This interpretation is supported by Pedersen et al. \cite{pedersen2010modeling} where the correlation between player emotions and level characteristics in a Super Mario Bros is investigated.
Here, the biggest factor for feeling challenged was the completion rate of the levels or similar aspects of failure, such as number of deaths.

In this work we adopt a similar probabilistic definition: the difficulty of a level is given by the win probability, which empirically is the completion rate and can be computed as the number of times a given level has been completed over the total number of attempts on said level.
However, while this aggregated description of difficulty as the completion rate is intuitive, it does not offer a deeper and actionable understanding of the problem, 
such as how imposing a time or action limit affects the completion rate or how close to finishing a player was.
The nature of such data is censored since we do not have information about the complete playthrough, so to draw inspiration on how to deal with that, we can look to survival analysis \cite{lee2003statisticalsurvival}.

Survival analysis is branch of statistics that focuses on estimating unseen, or censored, data and is commonly used to estimate a time until an event.
There are multiple examples of using this approach to describe player behaviour using parametric distributions: Feng et al. \cite{feng2005traffic} used a generalised Weibull distribution to model online session length, and Bauckhage et al. \cite{Bauckhage2012HowTimes} tested various distributions, including a Weibull and Poisson-Gamma distribution, to estimate time until people lost interest in a game.
A survival analysis approach has been used to describe gameplay related behaviour in \cite{Isaksen2018ExploringAnalysis}, in which the authors investigated the operationalisation of perceived difficulty of levels in the game Flappy Bird.
By using player and playtest AI data, they computed an empirical survival function, $S(x)$, which describes the distribution of attempts that reached a given length in a level.
From this, the hazard function could then be used as an indicator of perceived difficulty.

The work presented in this article shares its nature with these last studies, in that we attempt to operationalise and abstract aspect of gameplay -- i.e. level difficulty -- using a parametric statistical distribution. The key points of departure are, that the model presented in this article is both built and evaluated on a large dataset of real player gameplay data; furthermore, we present a general framework to describe the operationalisation of difficulty, identify the appropriate distribution and evaluate its effectiveness.







\begin{figure}[t]
    \centering
    \setlength{\fboxsep}{0pt}%
    \setlength{\fboxrule}{1pt}%
    \fbox{\includegraphics[width=0.90\columnwidth]{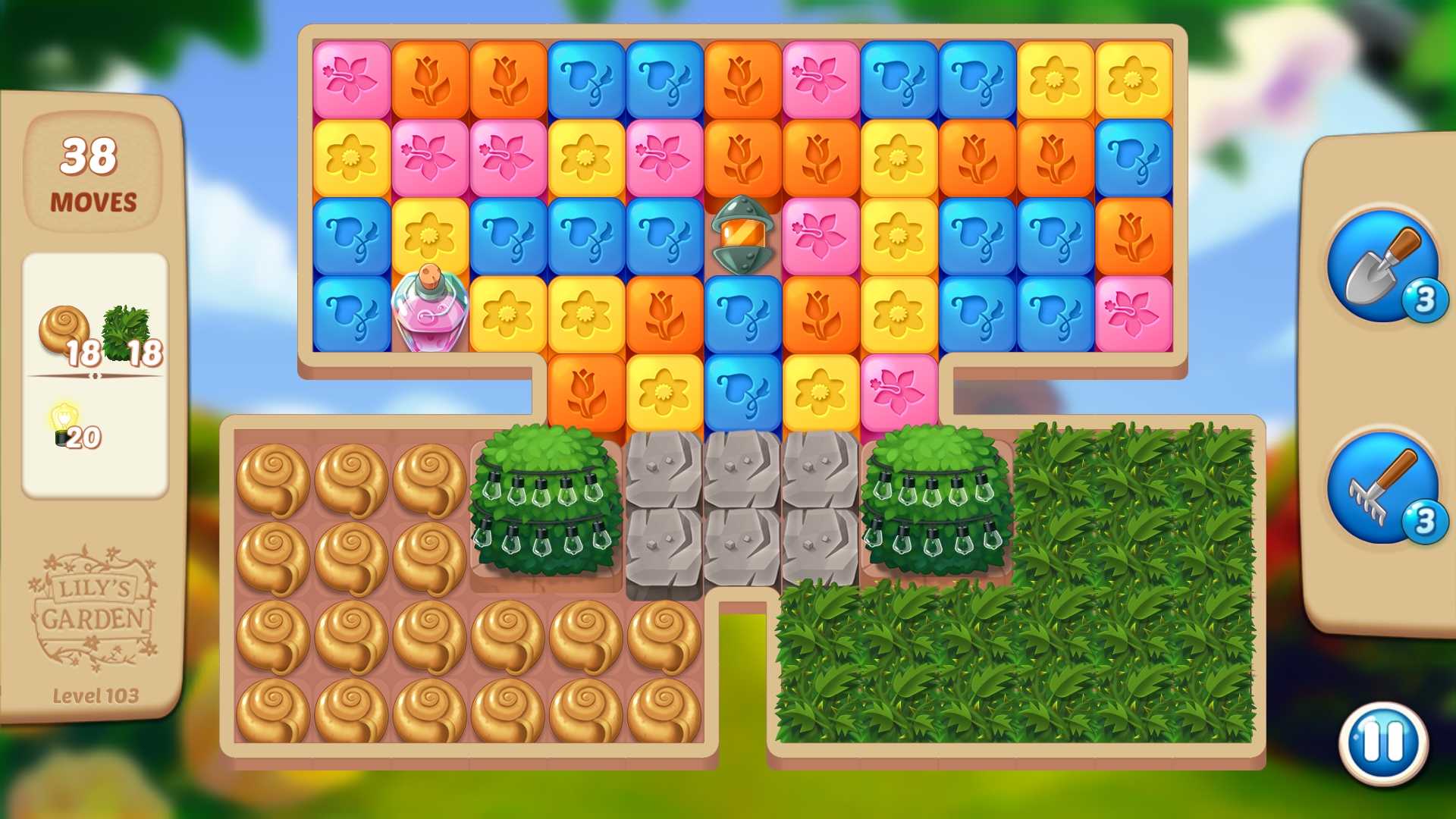}}
    \caption{An example of a level in Lily's Garden. The level goals are specified on the left side, and in-game boosters on the right side. These in-game boosters are very strong boosters that allow the player to complete the level more easily. 
    }
    \label{fig:example level}
\end{figure}

\section{Methods}

Let us start this section by briefly describing the puzzle game mechanics. Each level $\ell$ requires the player to collect a series of goals within a predetermined maximum number of actions, or moves, $M_\ell$. Each move consists of collapsing groups of adjacent board pieces by tapping on one of them. 
Creating more powerful board pieces that clear a large area of the board is possible by matching groups of at least 5 board pieces at the same time. An example of a level is shown in Fig. \ref{fig:example level}.

If the player completes all of the level goals with no more than $M$ moves, then we say that \textit{the attempt was successful}, and the player passes to the next level. Consequently, each player can complete each level at most once. Now, if the player consumes all of the permitted number of moves $M$ without completing the all of the level goals, then we say that \textit{the attempt was a failure}. In this case the player can either spend a virtual currency to obtain some extra moves (\emph{e.g.}, $+5$), or can decide to have one more attempt at the cost of a life. These lives regenerate automatically over time, and typically each player can get up to $5$ of lives at any given time.

For this study we use data sample from $L=4000$ levels which has been collected between 2020-06-01 and 2021-01-01.
For each level, the available data for each attempt consist of the number of moves used and whether the attempt was successful or not.
An initial data cleaning step is performed by excluding all incomplete attempts, \emph{i.e.}, attempts which are terminated prematurely either due to a technical issue in the game, or simply because the player deliberately quits the game. We also exclude attempts using special in-game boosters which usually inflate the number of attempts finishing within $k=0,1,2$ moves from the moves limit $M_\ell$.
The final input dataset consist of the frequency of moves used to complete the level (see Fig. \ref{fig:example move distribution}) and the overall completion rate, which is defined as the percentage of successful attempts over the total number of attempts, with an average of 350,000 successful attempts per level.





\begin{figure}[t]
    \centering
    \includegraphics[width=0.85\columnwidth]{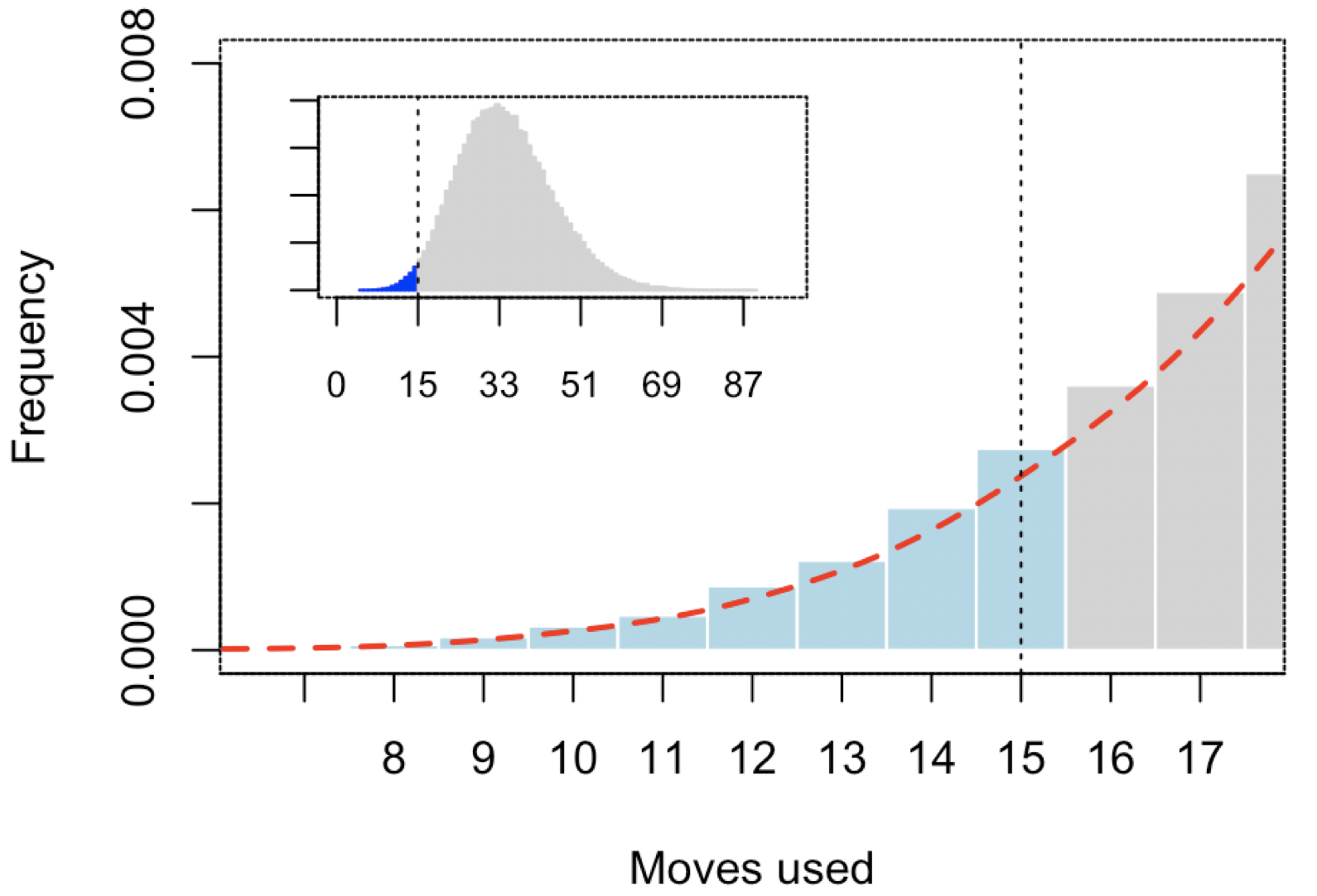}
    \caption{Illustration of the observed frequencies of moves to complete a level. The vertical dotted line indicates that moves limit is set to $M_\ell = 15$. The fitted curve is marked with a dashed line. The subplot in the top-left suggests the almost linear growth in the observed frequencies leads to fitting left tail of the negative binomial distribution.}
    \label{fig:Motivation Condition 2}
\end{figure}

The goal of the method is identifying a parametric distribution which can fit the number of moves used to complete a level to a \textit{good degree}, \emph{i.e.}, up the truncation point imposed by the moves limit $M_{\ell}^{*}$.
The fitted curve should match the observed frequencies, and the area under this curve should match the observed completion rate. As an illustration, Fig. \ref{fig:Motivation Condition 2} depicts the undesired situation where the fitted distribution is able to describe well the observed frequencies, but fails at matching the completion rate. We can expect this to occur for instance when the steady growth of the observed frequencies is almost linear and thus calibrated as the left tail of the distribution. These ideas are formalised below.

\begin{remark}
Let us note here that for other types of games, the definition of the input dataset would be analogous, for instance, by interchanging the role of moves used to complete the level by the units of time taken to complete the task. 
\end{remark}

\subsection{Calibration of model parameters}\label{sec:validation}

Given a level $\ell$ with a move limit $M_{\ell}^{*}$, let us denote
by $\hat{F}_{\ell}$ the empirical distribution of moves used to complete
the level. Let $\hat{c}_{\ell}$ be the observed level's completion
rate, \emph{i.e.}, the percentage of attempts that complete the level
within a maximum of $M_{\ell}^{*}$ moves. As depicted in Fig. \ref{fig:example move distribution} the
empirical move distribution $\hat{F}_{\ell}$ is truncated on the
right by $M_{\ell}^{*}$, but we assume that this data corresponds
to a censored observation of an underlying non-truncated distribution
$F_{\ell}$. Let us assume that $\hat{F}_{\ell}$ and $F_{\ell}$ have probability density functions, and denote them by $\hat{f}_{\ell}$ and $f_{\ell}$, respectively.

In these terms, our goal is to find a parametric model for the distribution
$F_{\ell}$, in such a way that following two conditions are met:

\begin{condition}
The fitted distribution, $F_\ell$, follows \emph{closely}
the empirical distribution, $\hat{F}_\ell$, all across the range $(0,M_{\ell}^{*}]$.
\end{condition}

\begin{condition}
The quantity $F_{\ell}(M_{\ell}^{*})$
approximates the observed completion rate $\hat{c}_{\ell}$.
\end{condition}

In this article we consider the Condition 2 as a validation step only;
that is to say, we do not explicitly enforce this condition as part
of the calibration algorithm. The rationale behind decision is that we aim at
establishing here a baseline for how much can be explained by focusing
only on fitting the truncated data. In other words, we are assessing
the degree in which Condition 1 can ensure that Condition 2 is fulfilled
as well.

Let us now describe calibration strategy for the model parameters. Given a parametric model
for the distribution $F_{\ell}$, we obtain the corresponding parameter
set $\text{\ensuremath{\theta}}_{\ell}$ by applying a Non-Linear
Least Squares (NLLS) regression over the range $(0,M_{\ell}^{*}]$,
which is were we can fully observe $\hat{F}_{\ell}$. Such a method
requires an initial guess $\text{\ensuremath{\theta}}_{0}$ of $\text{\ensuremath{\theta}}_{\ell}$
as an input, which, if incorrectly chosen, may
lead to a false negative due to a sub-optimal fit. In order to minimise
this risk we choose the initial guess by solving the following optimisation
problem:

\[
\theta_{\text{\ensuremath{0}}}^{*}(\ell):=\underset{\theta_{0}\in\Theta_\ell}{\text{arg min }}D(\hat{f_{\ell}},f_{\ell}^{(\theta_{0})}),
\]

where $\Theta_\ell$ denotes the search space for the initial guess $\text{\ensuremath{\theta}}_{0}$;
$f_{\ell}^{(\theta_{0})}$ is the distribution we get from NLLS by
using the initial guess $\text{\ensuremath{\theta}}_{0}$; and $D$
is a distance between the distributions $\hat{F}_{\ell}$
and $F_{\ell}^{(\theta_{0})}$ over the range $(0, M_\ell]$. Here we shall use the Kolmogorov-Smirnov
distance (see \cite{Kolmogorov-SmirnovMathematics}) which in this case is simply given by
\begin{align}
D(\hat{f_{\ell}},f_{\ell}^{(\theta_{0})}) & :=\max_{m\leq M_{\ell}^{*}}\left|\hat{F}_{\ell}(m)-F_{\ell}^{(\theta_{0})}(m)\right|\nonumber\\
 & =\max_{m\leq M_{\ell}^{*}}\left|\sum_{m'\leq m}\left(\hat{f}_{\ell}(m')-f_{\ell}^{(\theta_{0})}(m')\right)\right|.\label{eq:KS D}
\end{align}

Notice that in these terms Condition 1 can be rewritten as $D(\hat{f_{\ell}},f_{\ell}^{(\theta_{0}^{*}(\ell))})< \delta$,
for a small enough $\delta$, say $5\%$.

\subsection{Requirements for the underlying parametric distribution}

\begin{figure}[t]
    \centering
    \includegraphics[width=0.90\columnwidth]{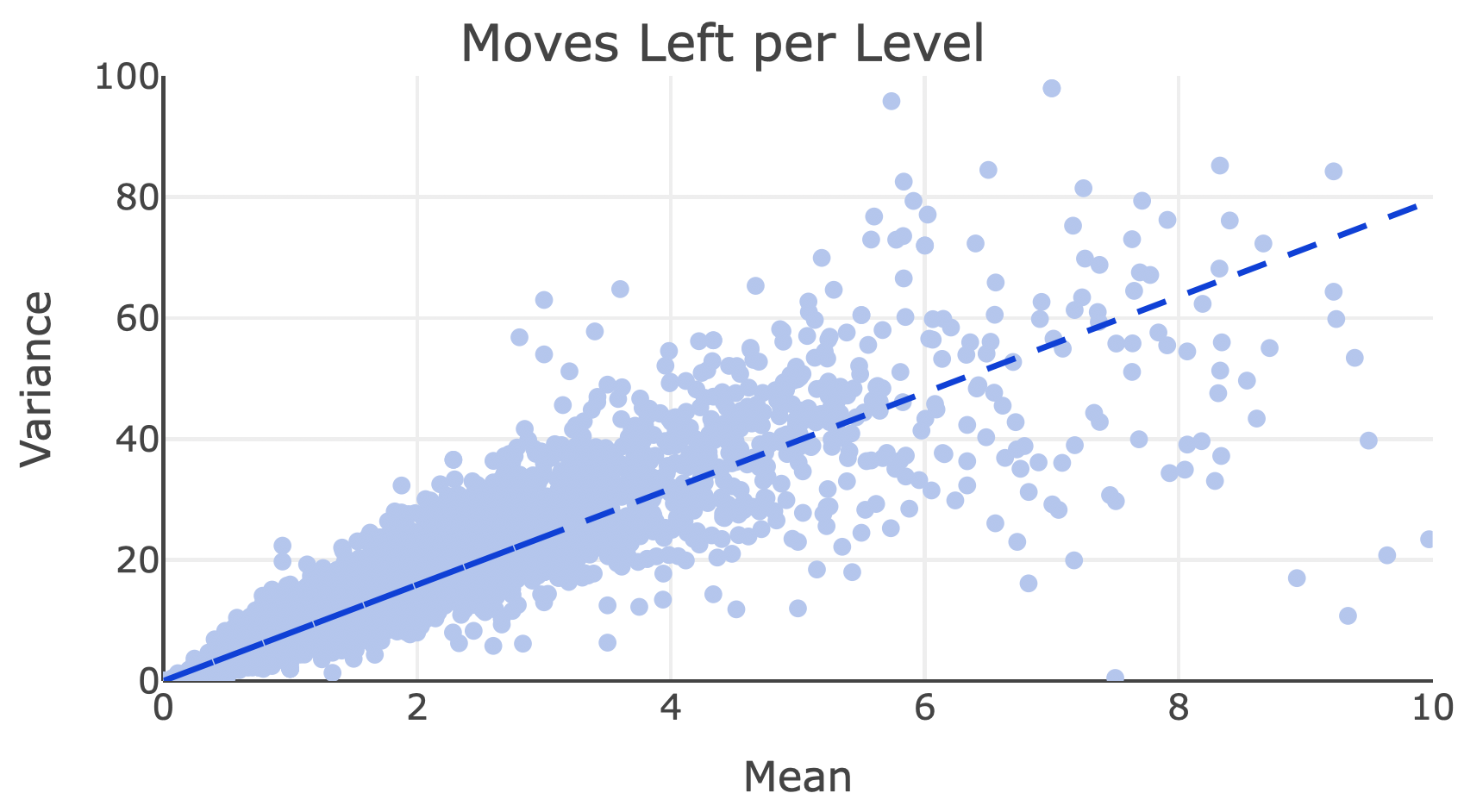}
    \caption{Illustration of the linear relationship between
the mean and the variance of number of moves left to complete level.
    }
    \label{fig:Motivation for the NB}
\end{figure}

Our target distribution (\emph{i.e.}, moves used to complete the level)
takes only non-negative integer values. Consequently, in order to
fit a parametric model we can use a non-negative integer-valued distribution
(\emph{e.g.}, negative binomial) or, alternatively, work with a discretization
of a non-negative continuous distribution (\emph{e.g.}, the gamma
distribution). 

In order to delimit the list of potential distributions we could use
for our analysis, we start by looking at the pattern depicted by
Fig. \ref{fig:Motivation for the NB} which suggests that there is a strong linear relationship between
the mean and the variance of number of moves left to complete level.
More precisely: Let $M_{\ell}(n,i)$ be the number of moves left at
the end of the $n$-th attempt of the $i$-th player to pass level
$\ell$. Each point of this graph corresponds to one of the levels
$\ell=1,...,L$ in our sample ($L = 4000$), and the coordinates $x$ and $y$ axis
equal the mean, $\mu_{\ell}$, and the variance, $\sigma_{\ell}$, of $M_{\ell}(n,i)$, respectively,
where $n$ and $i$ vary over all of the attempts that took place
during the observation period. The dashed line shows the result of
performing a linear regression of $\sigma_{\ell}^{2}$ with respect
to $\mu_{\ell}$ with no intercept -- $i.e.$, we consider a model of
the form $\sigma_{\ell}^{2}\approx\psi\mu_{\ell}$. The goodness of
this fit (\emph{i.e.,} $R^{2}\approx85\%$, $\text{p-value}<10^{-16}$)
suggests the aforementioned strong linear relationship between the
mean $\mu_{\ell}$ and the variance $\sigma_{\ell}^{2}$ of $M_{\ell}$.
Further it implies a necessarily condition that our parametric model
for $M_{\ell}$ should satisfy.


\subsection{Negative binomial distribution as a baseline}
Based on the above, is clear that the most natural non-trivial starting point is to consider a negative binomial distribution since it is a well-known non-negative integer-valued distribution exhibiting a linear relation between its mean and variance:

\[
f_\ell(m) := \binom{m+n-1}{m}(1-p)^{n}p^m,\quad \text{for } m=0,1,2,...
\]

As for the search space for the initial guess we shall use $\Theta_\ell:=[1,10 M_\ell]\times[0.001, 0.999]$.

Two remarks are in order here: first of all, note that the negative binomial distribution is also referred to as the \textit{Poisson-gamma distribution} since it is equivalent to a Poisson distribution with intensity parameter $\lambda$
where the $\lambda$ itself is allowed to be random by following a gamma distribution. Second, a more sophisticated approach would be to work with a discretization of a Tweedie distribution for which it is well-known that $\sigma_{\ell}^{2}=\psi\mu_{\ell}^{p}$, or even a Poisson-Tweedie distribution for which $\sigma_{\ell}^{2}=\mu_{\ell}+\psi\mu_{\ell}^{p}$ \cite{Bonat2018ExtendedPoisson, jorgensen1997theory}. 
However, we let this investigation for future work since our initial
exploration (see Fig. \ref{fig:Motivation for the NB}) suggests that considering a \textit{dispersion parameter} of $p=1$ could provide
already a very good starting point.

\section{Results}

To estimate the validity of our approach, we tested it on 4000 levels from the puzzle game Lily's Garden: first, we analyse the overall results of fitted distribution parameters on all of the levels.
In a second step, based the conditions described in the previous section regarding the fitted distributions, we discuss the goodness of the fit of the resulting model, thus evaluating the ability of the model to describe the player behaviour.
Lastly, we validate whether the model is able to describe the levels' canonical definition of difficulty -- \emph{i.e.} completion probability -- as well as the aforementioned behaviour.


\subsection{Distribution parameters}


\begin{figure}[t]
    \centering
    \includegraphics[width=0.90\columnwidth]{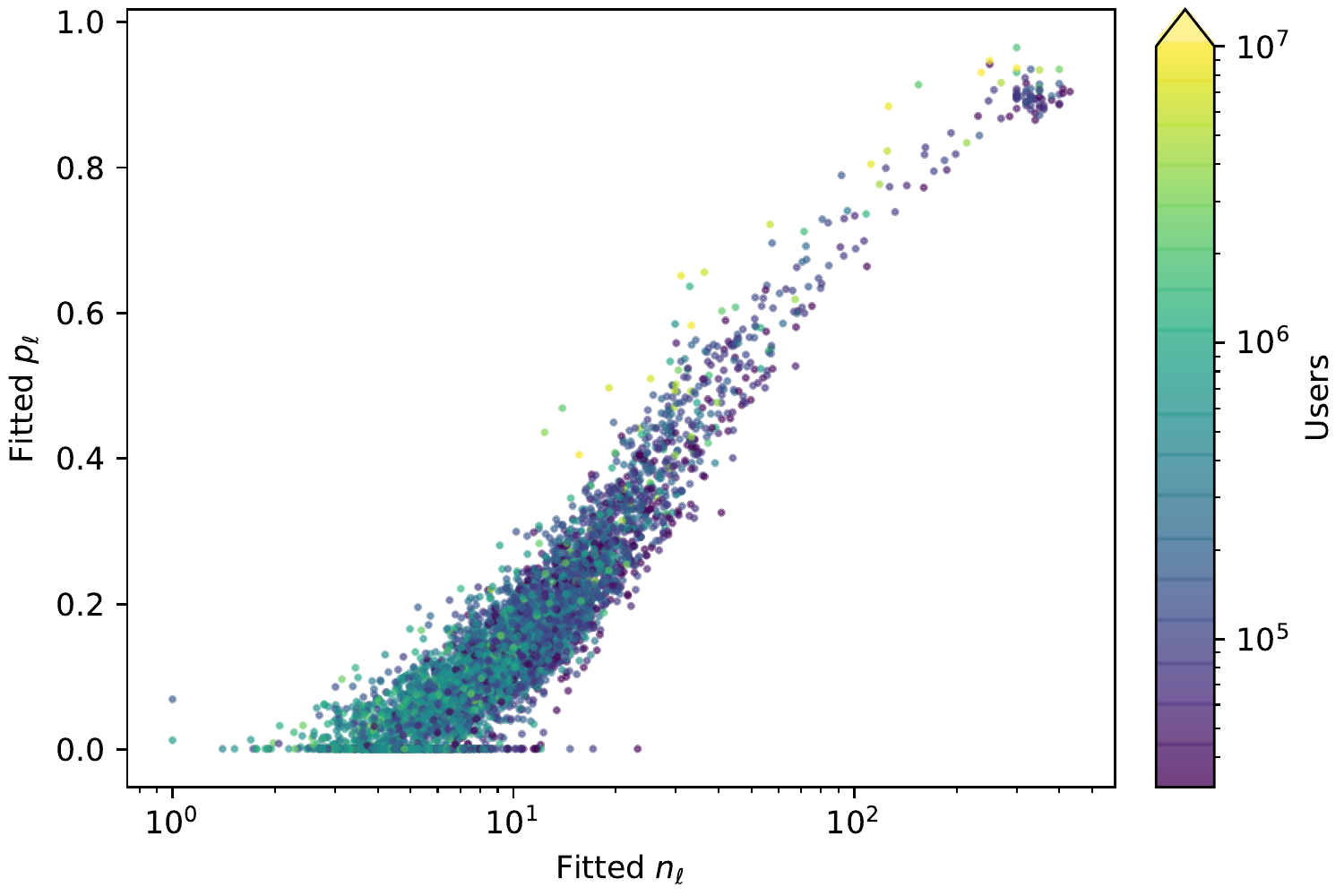}
    \caption{Log-linear plot of the fitted parameters $p$ and $n$ for each level.
    The color indicates the number of users that have played the given level. 
    }
    \label{fig:p versus n}
\end{figure}

Figure \ref{fig:p versus n} shows the fitted parameters obtained from the execution of the algorithm on $L=4000$ levels from the puzzle game Lily's Garden. Each of these points represent the parameters $(n_\ell,p_\ell)$ of a negative binomial model fitted to the distribution of moves used to complete each level $\ell=1,2,...,L$. It can be seen that the majority (\emph{i.e.}, $83\%$) of the levels fall within a central cluster defined by $0.001<p_\ell\leq 1$ and $1 \leq n_\ell \leq200$. For this central cluster it is apparent that the parameters $(n_\ell,p_\ell)$ follow a log-linear relationship $\log(n_\ell)=a p_\ell+b$ relationship ($R^2=87\%$), where $a$ and $b$ are global constants not depending on the level. 
This indicates that the level's move distribution can possibly be driven by a single parameter, which would enable level designers to easily compare levels to one another.

For this purpose, the so-called \textit{scale parameter} ($\vartheta_\ell$) could be considered, which describes the spread of a distribution -- \emph{i.e.}, the larger the scale parameter, the more spread out the distribution. This numerical parameter is often considered in the context of a parametric family of probability distributions, and in the case of negative binomial distributions it is given by this simple expression
\[
\vartheta_\ell := \frac{1-p_\ell}{p_\ell}.
\]
Notice that from this expression we can derive the $(n_\ell, p_\ell)$ as 
\[
p_\ell = \frac{1}{1+\vartheta_\ell},\;\text{ and }\;
n_\ell = \exp\left(a\left(\frac{1}{1+\vartheta_\ell}\right)+b\right).
\]

There are also two other notable clusters. The first of these clusters is defined by $p_\ell=0.001$ and consists of $15\%$ of sampled levels. The common feature for all instances in this cluster it that the parameter fitting threshold had been reached, which will be explored in more detail in Section \ref{subsec:filtering}. The second cluster is defined by $n_\ell>200$ and consists of $2\%$ of the levels in our sample. Inspecting the instances in this -- high $n_\ell$, high $p_\ell$ -- cluster, 
we encountered either tutorial levels or levels with a specific type of game mechanic that channels the players to rather restrictive type of game play.


It is worth noting that, by design, the tutorial levels tend to exhibit a lower variance than the rest of the levels, either by fixing the random seed or overall layout and ideal strategy of the level.
This reduced dependence on randomness may therefore also lead to a move distribution with smaller variance. In the same manner, we have observed that the levels containing the channelling mechanic that restricts gameplay lead to a less random play experience. Such information may be particularly useful to level designers since creating levels where the chance of winning is completely determined by chance removes any agency from players and are potentially not very fun to play. Being able to identify such levels can therefore provide a more quantitative measure of level randomness.

\begin{figure}[t]
   \centering
   \includegraphics[trim={0cm 0.0cm 0cm 0cm}, width=0.90\columnwidth]{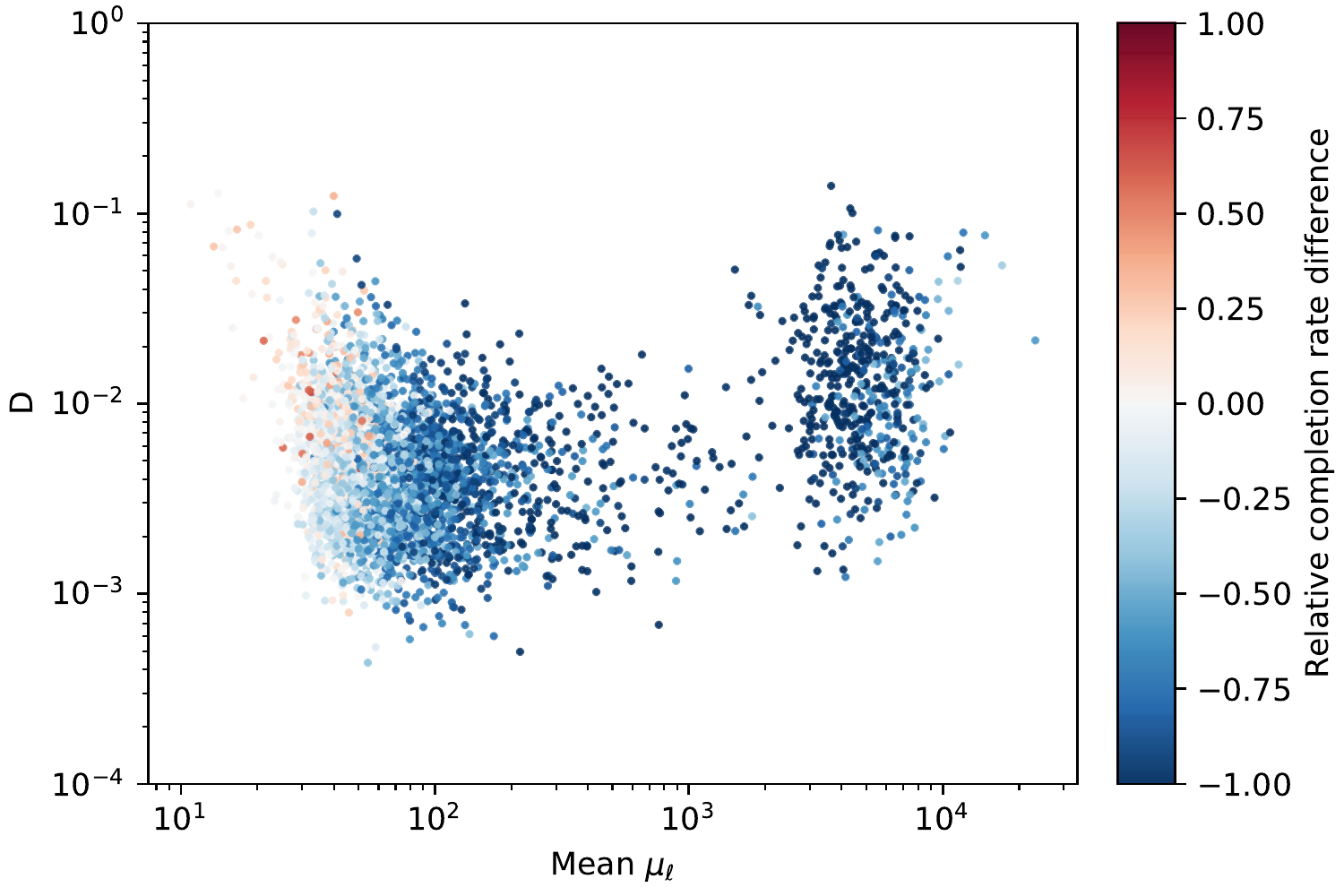}
   \caption{Log-log plot of the Kolmogorov-Smirnov test statistic $D$ and the mean of the fitted distributions. The colours show the relative difference between the expected and actual completion rate. 
   }
   \label{fig:D versus mean with cr color}
\end{figure}

\subsection{Condition 1 and validity of fits}
\label{subsec:filtering}

The initial condition laid out in Section \ref{sec:validation} states that the fitted distribution $F_\ell$ should closely follow the empirical distribution $\hat{F}_\ell$. To determine whether this is true, we use the Kolmogorov-Smirnov distance $D$ as defined by Eq. (\ref{eq:KS D}). 
To give an overview of the link between the distribution parameters and $D$, Fig. \ref{fig:D versus mean with cr color} plots $D$ against the mean ($\mu_\ell=n_\ell \frac{1-p_\ell}{p_\ell}$) of the fitted distribution, and coloured by the relative difference $(c_\ell-\hat{c_\ell})/\hat{c_\ell}$.
What we find is that $99\%$ of the levels satisfy $D<5\%$, meaning that the fitted distributions describe the empirical data very well in many cases and thus fulfil \textbf{Condition 1}.

One thing to note is that in some cases, the parameter boundaries were reached during the fitting process.
This was observed to happen in around $15\%$ of the levels and typically lead to $p_\ell=0.001$.
These levels appear in the right-most cluster in Fig. \ref{fig:D versus mean with cr color} and are defined by $\mu_\ell>10^3$.
This was typically observed to happen when the empirical move distribution only exhibited a steadily increasing trend, leading to instances where only using the tail of the distribution would best describe this simple behaviour.
We consider those examples bad fits due to the method not converging and exclude them for the rest of the analysis in the next section.

Before moving on to the next part of the analysis, we first attempt to isolate what differentiates the levels that show a good fit from the other ones.
Specifically, we first investigate whether different game mechanics influence the move distribution.
For this purpose, we use a logistic regression to model whether the level fit converged or not in order to estimate the impact of specific board pieces.
The results indicate that timing mechanics generally lead to a better fit while one specific spawning mechanic (\emph{i.e.}, the collect goals first appear after interacting with the spawner) lead to a worse fit.

One thing that is worth noting is that the data used for this analysis disregarded attempts that used various in-game help items (\emph{i.e.}, extra moves, boosters, etc).
If a player finds a level to be difficult or frustrating, subsequent attempts by the player may be disregarded because they use helping items, distorting the move distribution.
A number of observations support this hypothesis: When only considering the move distribution of the second attempt of players, the fraction of levels that successfully converged increased by about $+5\%$.
Additionally, the fraction of attempts in which players used in-game boosters and help items were up to $+18\%$ more frequent in non-converging examples than convergent ones; thus, more attempts are ignored on average for non-converging examples. In our data processing step, these attempts were filtered out because they exhibited a clear artificial alteration of the curve, especially in the last two moves of the levels. Part of the explanation for the divergent fits can therefore also be related to the data.


\begin{figure}[t]
    \centering
    \includegraphics[width=0.95\columnwidth]{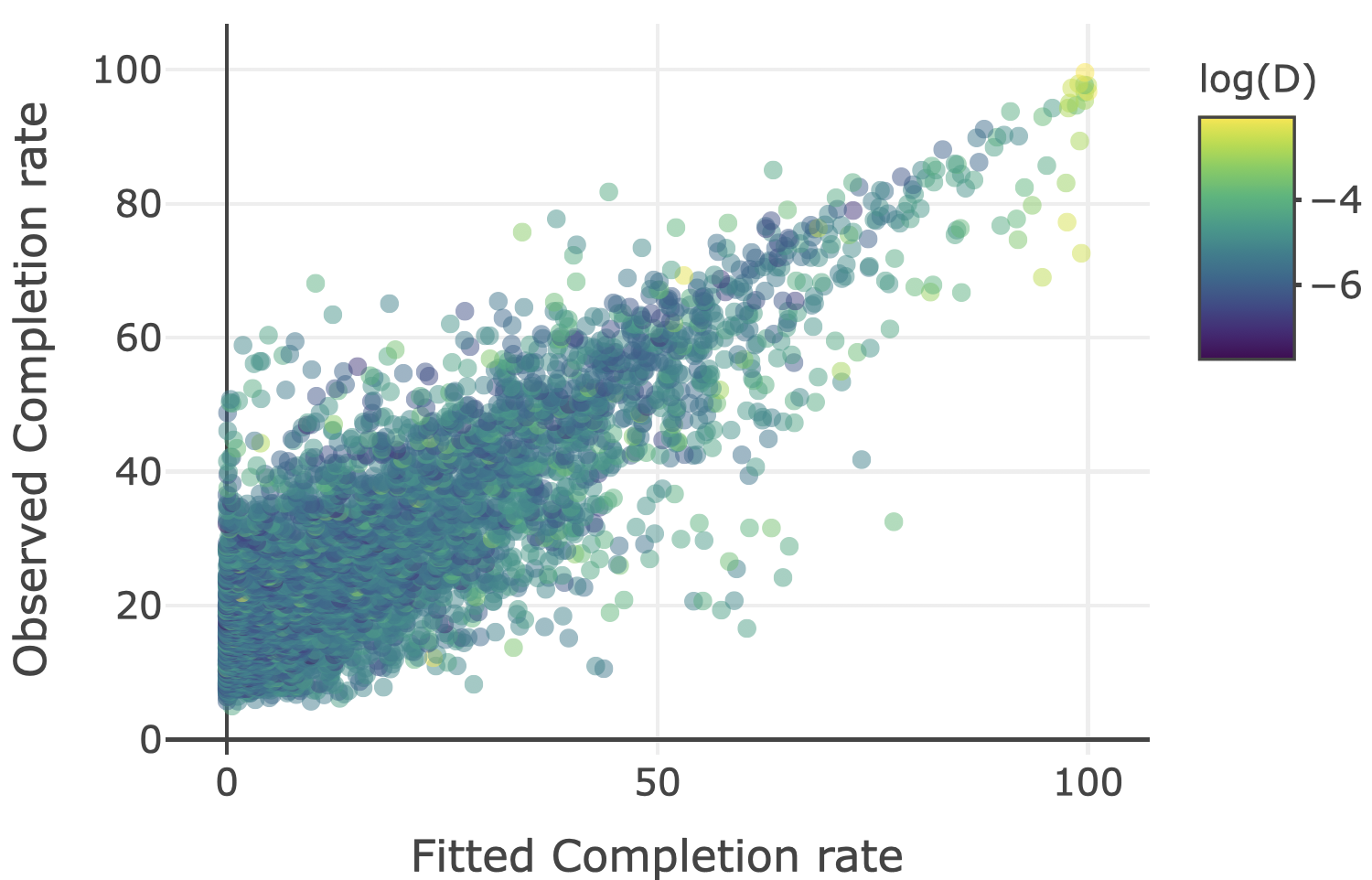}
    \caption{Comparison between the observed completion rates and the fits obtained from the calibration algorithm.
    }
    \label{fig:completion rate comparison}
\end{figure}


\subsection{Condition 2: completion rate comparison}\label{sec:completionrate}

\begin{figure*}[ht]
    \centering
    \includegraphics[width=1.75\columnwidth]{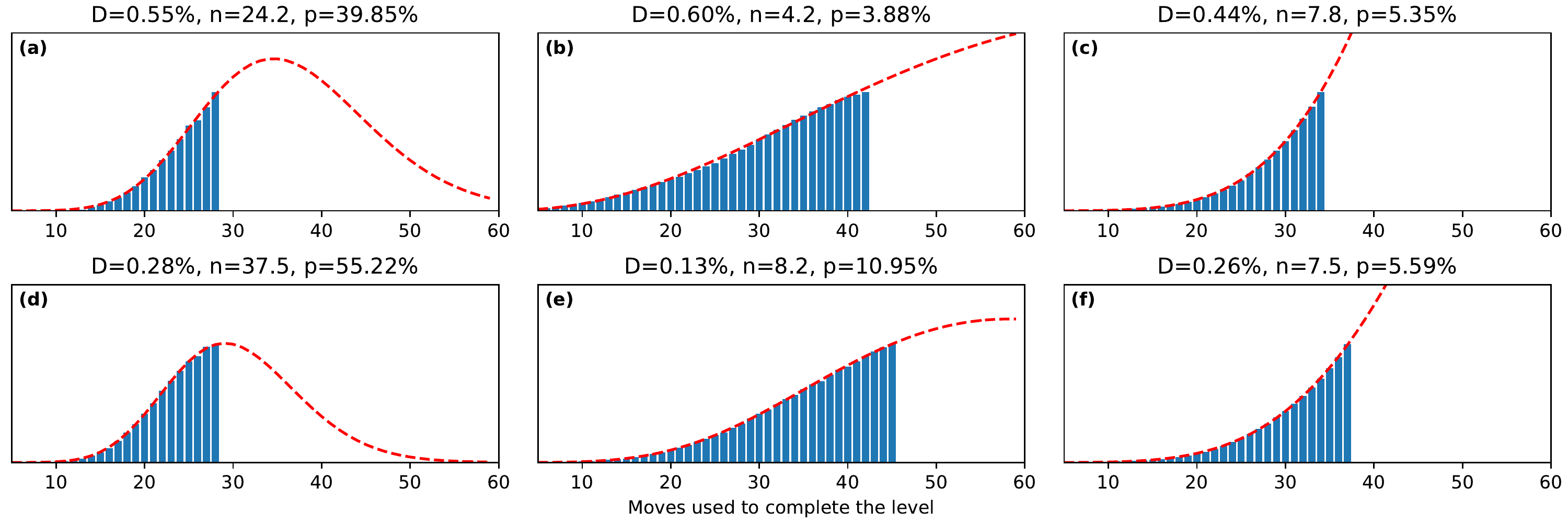}
    \caption{Subplots in the top (\textbf{a-c}) and bottom rows (\textbf{d-f}) correspond to instances when the observed completion rate $\hat{c}_\ell\approx20\%$ and $\hat{c}_\ell\approx40\%$, respectively. The first, second and third columns exemplify cases we found a good (\textbf{a} and \textbf{d}), medium (\textbf{b} and \textbf{e}) and low (\textbf{c} and \textbf{f}) agreement between the observed and fitted completion rates, respectively.
    }
    \label{fig:6 example levels}
\end{figure*}
The second condition states that the expected completion rate, $\hat{F_\ell}(M_\ell^{*})$, should approximate the observed completion rate, $\hat{c_\ell}$. In order to assess this condition, we first notice that the two values are strongly correlated as exhibited by their Pearson's correlation coefficient of $\rho = 83\%$. Further, Fig. \ref{fig:completion rate comparison} suggests that the observed and fitted completion rates are related to each other by means of the linear relationship
\begin{equation}
c_\ell \approx 1.035\hat{c}_\ell - 0.104\label{eq: final lm}    
\end{equation}
with an adjusted coefficient of determination of $R^2=75\%$. Equation (\ref{eq: final lm}) suggests that the completion rates tend to be underestimated, especially at low completion rates (\emph{i.e.}, for very hard levels where the average player will need the equivalent of 8 or more attempts are needed to complete the level). Based on these arguments we can consider the \textbf{Condition 2} has been met as well. 

In practice level designers typically work with ranges of the completion  rate rather than point estimates, so that they can classify the levels in  classes (\emph{e.g.}, ”easy”, ”very hard”). Consequently, the current results are positive and very promising. One could however also look at point estimates of the completion rates, for instance under the light of the \textit{absolute percentage error} given by 
$\varepsilon_\ell:=\left | c_\ell/\hat{c}_\ell - 1 \right|$. By doing this, we have observed that the median value of $\varepsilon_\ell$ revolves around the $49\%$, and it goes down to $23\%$ when adjusting according to Equation (\ref{eq: final lm}). 

In order to get a better understanding on what leads to the aforementioned underestimation (\emph{i.e.}, cases where $c_\ell < \hat{c_\ell}$), we exemplify in Fig. \ref{fig:6 example levels} what happens in a series of scenarios: Scenario 1, as illustrated by subplots \textbf{a} and \textbf{d}, corresponds to cases where the relative error between $\hat{c}_\ell$ and $c_\ell$ is small. Scenario 2 in subplots \textbf{b} and \textbf{e} show cases where error is medium. And finally Scenario 3 in subplots \textbf{c} and \textbf{f}  corresponds cases with a major underestimation. In Scenario 3, it can be seen that it is only the tail of the distribution that is used to describe the data. 
A similar phenomenon was also observed in the cases where the fitting method did not converge: Due to the available player data and steady increase in completions, only the tail is required to describe this relatively simple behaviour.
However, contrary to those cases, these levels are in more of a continuum: It is more likely to underestimate at low completion rates where more data is censored, while for higher values of $\hat{c}_\ell$ (like in Scenario 1 and Scenario 2) we have more information about the distribution is available which further constrains $f_\ell$.

In order to see if there are any specific game mechanics that may cause a difference between the completion rates, a similar method as section \ref{subsec:filtering} is used.
Instead of using a logistic regression for predicting whether it was a good fit or not, a linear regression is used to predict the difference between expected and actual completion rate.
The results are similar to the findings in the previous section regarding successful fitting: Levels with timing or other gameplay restrictive mechanics lead to a higher expected completion rate.
Interestingly, board pieces with colour-matching mechanics tend to lead to too low expected completion rates.
A way to possibly interpret this is that goals which can be completed at a steady pace (such as colour mechanics) lead to a more steadily increasing ramp-like distribution, leading to completely underestimating the completion rate due to more degrees of freedom in the fitting.
Timing mechanics, on the other hand, may require more planning that appear as a more constrained minimum number of moves spent which leads more defined distribution around a given move and less variance that may be detrimental to the modelling method.
That said, there are additional factors not considered (such as level topology), so more work is required to establish any link between the completion rate difference and game mechanic.


\begin{figure}[t]
    \centering
    \includegraphics[width=0.85\columnwidth]{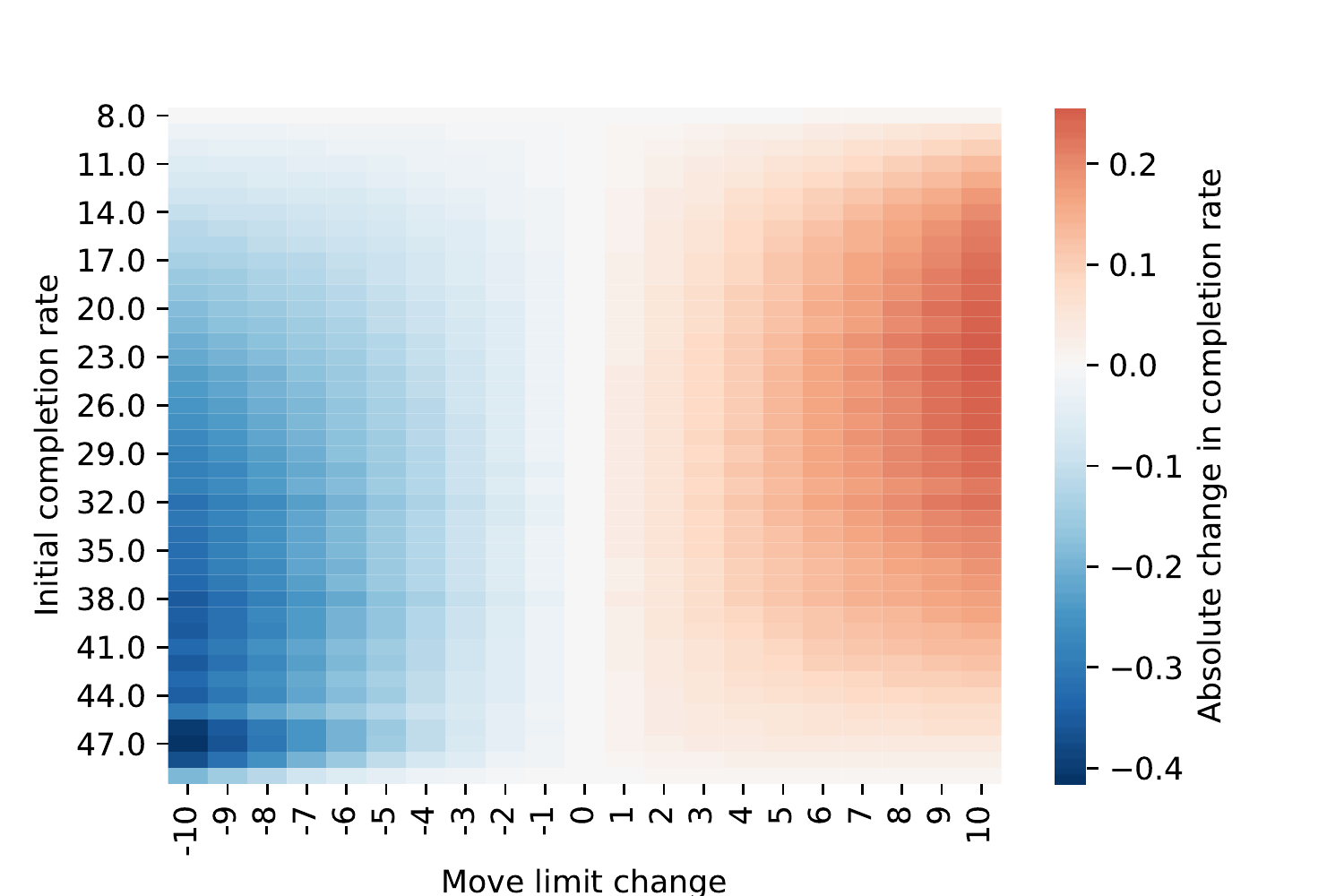}
    \caption{2D plot of how the completion rate is expected to change depending on the initial completion rate and the change in moves limit. The initial completion rates are binned in bins spanning 2\%, and the new expected completion rate is adjusted by the trend from Fig. \ref{fig:completion rate comparison}.
    }
    \label{fig:movelimit change comparison}
\end{figure}

\section{Discussion and future work}

In 85\% of the levels we are able to find a negative binomial distribution that describes the player data well.
Additionally, we are able to derive estimations of different game play features, such as level randomness and board piece descriptors, that can give additional insights to the game designers.
That said, there are still some open questions to address about the current approach related to the modelling and possible use-cases, which will be discussed in this section.

\subsection{Changing the move limit}

One of the discussed use-cases of modelling these distributions is that level designers can estimate what changing the move limit would mean for the completion rate.
To examine how the completion rate is affected by changes in the move limit, Fig. \ref{fig:movelimit change comparison} shows how the predicted absolute change in expected completion rate depends on the initial completion rate and change in move limit.
As a rule of thumb, the completion rate seems to change on average by 2\% with slightly lower sensitivity at high or low completion rates.
From discussions with level designers, this is consistent with their commonly used heuristic.

Another insight is that adding or removing moves is an asymmetric operation, where the rate of change is bigger when removing moves.
While this is also expected since the negative binomial distribution itself can be asymmetric and can potentially have a long right tail (and thus less sensitive to adding moves), it suggests that game designers need to be more careful when removing time or actions to increase difficulty because of this asymmetric change.

One possible limitation of this argument is that it assumes that the distribution parameters will stay the same if the move limit changes.
However, this is not necessarily true since players may change their behaviour when closer to the move limit.
For instance, a common strategy is to set up powerful board piece combinations and fire them off in the end to maximise the score (regardless of whether there is an explicit score or not).

\subsection{Player skill}

The perceived level difficulty depends not just on level randomness but also the player skill.
So far the level randomness was linked to the variance of the fitted distributions, but logically the move distributions should also be affected by the skills that have played the level.
Indeed, this phenomenon is something that level designers experience in their day to day work: As more players reach older levels, the completion rate slowly changes, which makes it necessary to have a constant maintenance of all levels.

As a next step, investigating how the level difficulty changes over time in a longitudinal study using different player cohorts may provide meaningful insights on player skill and also model how this affects the distribution parameters.
This can then be used for a more proactive and automatic approach to difficulty adjustment that ensures a coherent play experience for both old and new users.

\subsection{Playtesting}

\begin{figure}[t]
    \centering
    \includegraphics[trim={0 0 0 -2.7cm},width=0.85\columnwidth]{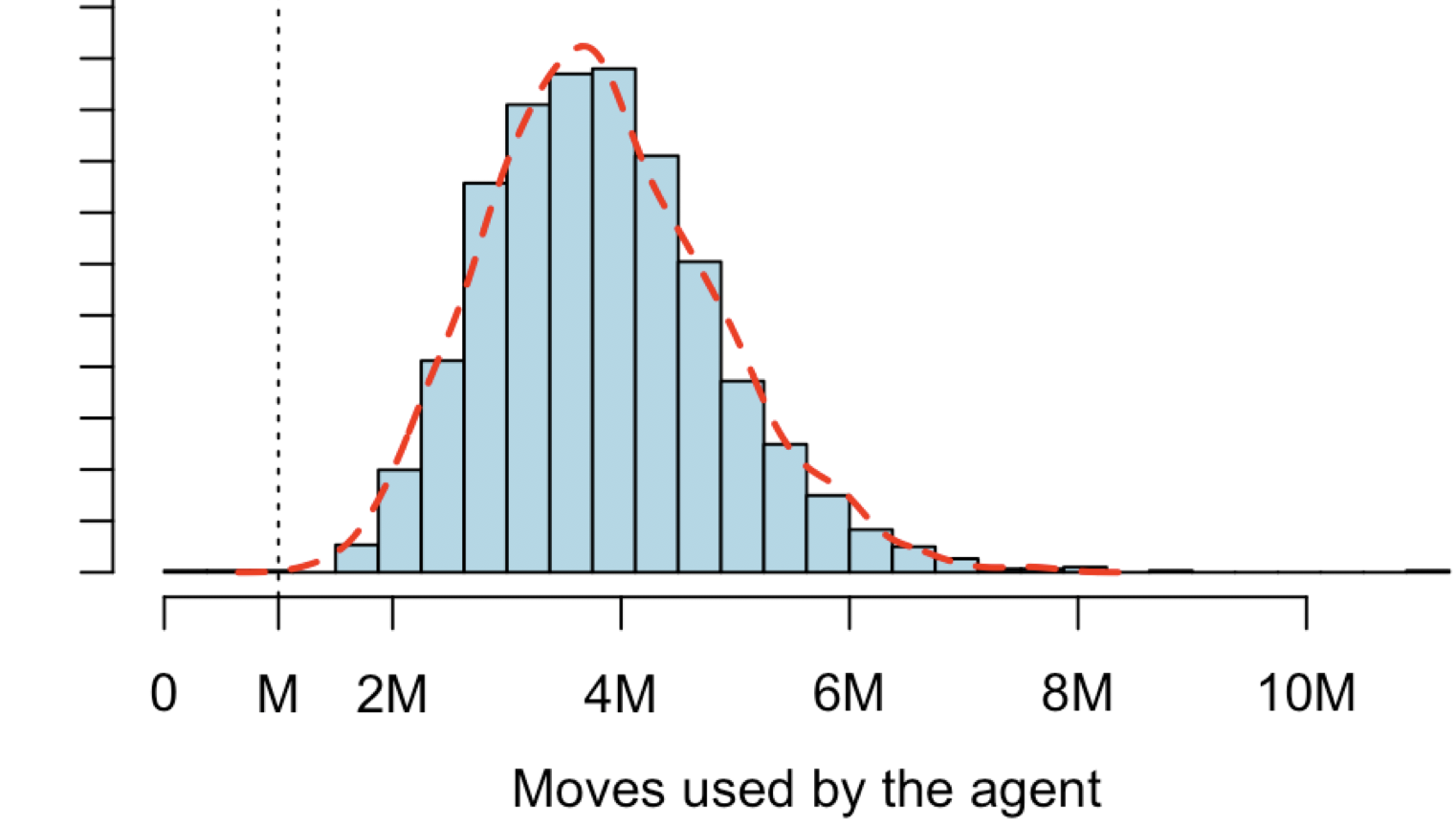}
    \caption{Histogram of the moves used by a certain AI agent to complete a specific Lily's Garden. The dashed line represents the negative binomial fit.}
    \label{fig:agent moves distr}
\end{figure}

Playtesting is crucial for game developers since this process provides a reliable way to identify bugs and potential design flaws in a safe environment before going to market. 
This process, however, tends to be so expensive and slow that game developers are increasingly starting to automate this by means of AI agents using,  for instance, reinforcement learning techniques (\emph{e.g.} \cite{kristensen2020strategies} and references therein). This context also provides an interesting set-up to gain deeper insight into the techniques derived in the present article and further potential applications. 

Indeed, given that playtest agents are trained in the same environment as human players, we can for instance let the agent play using all of the normal rules and game mechanics but without the constraint on the move limit, $M_\ell$. Figure \ref{fig:agent moves distr} shows the distribution of moves generated by one of the playtest agents considered in \cite{kristensen2020estimating} when testing a given level. This particular agent performed sub-par with respect to the average human player, but what is relevant is that we are able to visualise its whole move distribution even beyond the limit $M$.
We can thus fit our proposed negative binomial distribution across the whole $(0,10M]$ range, \emph{i.e.}, without truncation. 

In line with our expectations, we get a really good fit as described by a Kolmogorov-Smirnov distance of $D=1.8\%$. This sparks the following question regarding playtest agents: \textit{Can exhibiting a negative binomial distribution be regarded as a necessary condition to declare that the AI agent is playing in a human-like manner?}


Finally we highlight another connection with the results reported in \cite{kristensen2020estimating}. In that work the authors report that the $5\%$ best runs of the agent on a given level were the strongest predictor of the actual completion rate. Clearly this $5^{th}$ percentile is a quantity that can be derived explicitly as formula of the fitted negative binomial parameters. In this sense we can also study whether the fitting procedure proposed here can be further used as post-processing strategy to estimate completion rates from data generated by playtest agents with sub- or even super-human performance. 

\subsection{Other games}

In this work we have investigated the application of the proposed method to a mobile puzzle game; however, there is nothing in our assumptions that rules out that the same distribution can be used not just for similar puzzle games with discrete moves and action limit but also other genres such as platform or even competitive games.

Generally, puzzle games tend to be very focused on solving the level goal as fast as possible.
Although some games also provide a score, it is a limited number of factors (randomness and skill) that affect the distribution of moves.
However, in other game genres, there may be other factors and incentives for playing:
in platform games, players are encouraged to explore and test out different strategies, and in competitive games, players may want to beat their opponent as fast as possible, with randomness playing less of a role than the relative skill of players.
A promising venue for future research is therefore using this modelling approach across genres to test and validate its generalisability to different player behaviours.



\section{Conclusion}

In this research work we set out to determine a richer way of describing the level difficulty in puzzle games.
Specifically, we propose that the move frequency distribution of the players for completing a level follows a negative binomial.
Using data from $4000$ levels from the game Lily's Garden as a case study, the results showed that:
\begin{itemize}
    \item The negative binomial is able to describe the move distribution of around $85\%$ of the levels, and the method can easily be extended to other types of games.
    \item Describing the levels is possible using a single parameter -- that is the scale parameter $\vartheta$ -- that describes the spread of the distribution.
    \item This more detailed description of the difficulty enables: (\textit{i}) estimating the effect of changing the move limit; \\
    (\textit{ii}) estimating the level randomness; and \\
    (\textit{iii}) identifying deviations in player behaviour on a level.
\end{itemize}

In the remaining $\sim15\%$ of the cases where the method does not converge; the main issue is due to the data only exhibiting an increasing trend which leads to the method only using a very small part of the distribution to match it.
Similarly, the method also tends to underestimate the observed completion rate, $\hat{c_\ell}$, especially towards low completion rates.
A possible avenue for future research is therefore to extend on this model and include $\hat{c_\ell}$ as a parameter in the modelling rather than a constraint.
This has the promise of not only improving the predictions of the method but also ultimately enable estimating player skill and dynamically adjust difficulty to ensure an optimal player experience.



\section{Acknowledgements}

This work has been supported by the Innovation Fund Denmark and Tactile Games.
We also thank Arnau Escapa and Rasmus Berg Palm for fruitful discussions.

\bibliographystyle{plain}
\bibliography{referencesmanual,references}

\begin{thebibliography}{10}

\bibitem{Kolmogorov-SmirnovMathematics}
{Kolmogorov-Smirnov test - Encyclopedia of Mathematics}.

\bibitem{Bauckhage2012HowTimes}
Christian Bauckhage, Kristian Kersting, Rafet Sifa, Christian Thurau, Anders
  Drachen, and Alessandro Canossa.
\newblock {How players lose interest in playing a game: An empirical study
  based on distributions of total playing times}.
\newblock In {\em 2012 IEEE Conference on Computational Intelligence and Games,
  CIG 2012}, pages 139--146, 2012.

\bibitem{Bonat2018ExtendedPoisson}
Wagner~H. Bonat, Bent Jørgensen, Célestin~C. Kokonendji, John Hinde, and
  Clarice G.~B. Demétrio.
\newblock Extended poisson–tweedie: Properties and regression models for
  count data.
\newblock {\em Statistical Modelling}, 18(1):24--49, 2018.

\bibitem{csikszentmihalyi1990flow}
Mihaly Csikszentmihalyi and Mihaly Csikzentmihaly.
\newblock {\em Flow: The psychology of optimal experience}, volume 1990.
\newblock Harper \& Row New York, 1990.

\bibitem{Demediuk2017MonteAdjustment}
Simon Demediuk, Marco Tamassia, William~L. Raffe, Fabio Zambetta, Xiaodong Li,
  and Florian Mueller.
\newblock {Monte Carlo tree search based algorithms for dynamic difficulty
  adjustment}.
\newblock In {\em 2017 IEEE Conference on Computational Intelligence and Games,
  CIG 2017}, pages 53--59. Institute of Electrical and Electronics Engineers
  Inc., 10 2017.

\bibitem{DENISOVA2020102383}
Alena Denisova, Paul Cairns, Christian Guckelsberger, and David Zendle.
\newblock Measuring perceived challenge in digital games: Development \&
  validation of the challenge originating from recent gameplay interaction
  scale (corgis).
\newblock {\em International Journal of Human-Computer Studies}, 137:102383,
  2020.

\bibitem{Duque2020FindingTrial-and-Error}
Miguel~González Duque, Rasmus~Berg Palm, David Ha, and Sebastian Risi.
\newblock {Finding Game Levels with the Right Difficulty in a Few Trials
  through Intelligent Trial-and-Error}, 1 2020.

\bibitem{feng2005traffic}
Wu-chang Feng, Francis Chang, Wu-chi Feng, and Jonathan Walpole.
\newblock A traffic characterization of popular on-line games.
\newblock {\em IEEE/ACM Transactions On Networking}, 13(3):488--500, 2005.

\bibitem{gudmundsson2018human}
Stefan~Freyr Gudmundsson, Philipp Eisen, Erik Poromaa, Alex Nodet, Sami
  Purmonen, Bartlomiej Kozakowski, Richard Meurling, and Lele Cao.
\newblock Human-like playtesting with deep learning.
\newblock In {\em 2018 IEEE Conference on Computational Intelligence and Games
  (CIG)}, pages 1--8. IEEE, 2018.

\bibitem{Isaksen2018ExploringAnalysis}
Aaron Isaksen, Dan Gopstein, Julian Togelius, and Andy Nealen.
\newblock {Exploring Game Space of Minimal Action Games via Parameter Tuning
  and Survival Analysis}.
\newblock {\em IEEE TRANSACTIONS ON GAMES}, 10(2), 2018.

\bibitem{jorgensen1997theory}
B.~Jørgensen.
\newblock {\em The Theory of Dispersion Models}.
\newblock Chapman \& Hall/CRC Monographs on Statistics \& Applied Probability.
  Taylor \& Francis, 1997.

\bibitem{kristensen2020strategies}
Jeppe~Theiss Kristensen and Paolo Burelli.
\newblock Strategies for using proximal policy optimization in mobile puzzle
  games.
\newblock In {\em International Conference on the Foundations of Digital
  Games}, pages 1--10, 2020.

\bibitem{kristensen2020estimating}
Jeppe~Theiss Kristensen, Arturo Valdivia, and Paolo Burelli.
\newblock Estimating player completion rate in mobile puzzle games using
  reinforcement learning.
\newblock In {\em 2020 IEEE Conference on Games (CoG)}, pages 636--639. IEEE,
  2020.

\bibitem{lee2003statisticalsurvival}
Elisa~T Lee and John Wang.
\newblock {\em Statistical methods for survival data analysis}, volume 476.
\newblock John Wiley \& Sons, 2003.

\bibitem{Lomas2017IsDifficultyOverrated}
J.~Derek Lomas, Kenneth Koedinger, Nirmal Patel, Sharan Shodhan, Nikhil
  Poonwala, and Jodi~L. Forlizzi.
\newblock Is difficulty overrated? the effects of choice, novelty and suspense
  on intrinsic motivation in educational games.
\newblock In {\em Proceedings of the 2017 CHI Conference on Human Factors in
  Computing Systems}, CHI '17, page 1028–1039, New York, NY, USA, 2017.
  Association for Computing Machinery.

\bibitem{pedersen2010modeling}
Christopher Pedersen, Julian Togelius, and Georgios~N Yannakakis.
\newblock Modeling player experience for content creation.
\newblock {\em IEEE Transactions on Computational Intelligence and AI in
  Games}, 2(1):54--67, 2010.

\bibitem{xue2017dynamic}
Su~Xue, Meng Wu, John Kolen, Navid Aghdaie, and Kazi~A. Zaman.
\newblock Dynamic difficulty adjustment for maximized engagement in digital
  games.
\newblock In {\em Proceedings of the 26th International Conference on World
  Wide Web Companion}, WWW '17 Companion, page 465–471, Republic and Canton
  of Geneva, CHE, 2017. International World Wide Web Conferences Steering
  Committee.

\end{thebibliography}

\end{document}